\documentclass{article}

\usepackage{PRIMEarxiv}
\usepackage{authblk}

\usepackage[utf8]{inputenc} 
\usepackage[T1]{fontenc}    
\usepackage{hyperref}       
\usepackage{url}            
\usepackage{booktabs}       
\usepackage{amsfonts}       
\usepackage{amsmath}        
\usepackage{nicefrac}       
\usepackage{microtype}      
\usepackage{algorithmic}   
\usepackage{algorithm}     
\usepackage{lipsum}
\usepackage{multirow}
\usepackage{makecell}
\usepackage{xcolor}
\usepackage{amssymb}
\usepackage{pifont}
\usepackage{fancyhdr}       
\usepackage{enumitem}      
\usepackage{graphicx}       
\graphicspath{{media/}}     

\definecolor{DarkGreen}{rgb}{0.0, 0.5, 0.0}
\definecolor{DarkRed}{rgb}{0.8, 0.0, 0.0}

\newcommand{\cmark}{\textcolor{DarkGreen}{\checkmark}}
\newcommand{\xmark}{\textcolor{DarkRed}{\ding{55}}}

\title{MedDialogRubrics: A Comprehensive Benchmark and Evaluation Framework for Multi-turn Medical Consultations in Large Language Models}

\usepackage{authblk}

\author[1]{Lecheng Gong}
\author[1]{Weimin Fang}
\author[1]{Ting Yang}
\author[1]{Dongjie Tao}
\author[1]{Chunxiao Guo}
\author[1]{Peng Wei}
\author[1]{Bo Xie}
\author[1]{Jinqun Guan}
\author[1]{Zixiao Chen}
\author[1]{Fang Shi}
\author[1]{Jinjie Gu}
\author[1]{Junwei Liu}

\affil[1]{Ant Group
}

\begin{document}
\maketitle
\begin{abstract}
Medical conversational AI (AI) plays a pivotal role in the development of safer and more effective medical dialogue systems. However, existing benchmarks and evaluation frameworks for assessing the information-gathering and diagnostic reasoning abilities of medical large language models (LLMs) have not been rigorously evaluated. To address these gaps, we present \textbf{MedDialogRubrics}, a novel benchmark comprising 5,200 synthetically constructed patient cases and over 60,000 fine-grained evaluation rubrics generated by LLMs and subsequently refined by clinical experts, specifically designed to assess the multi-turn diagnostic capabilities of LLM. Our framework employs a multi-agent system to synthesize realistic patient records and chief complaints from underlying disease knowledge without accessing real-world electronic health records, thereby mitigating privacy and data-governance concerns. We design a robust Patient Agent that is limited to a set of atomic medical facts and augmented with a dynamic guidance mechanism that continuously detects and corrects hallucinations throughout the dialogue, ensuring internal coherence and clinical plausibility of the simulated cases. Furthermore, we propose a structured LLM-based and expert-annotated rubric-generation pipeline that retrieves Evidence-Based Medicine (EBM) guidelines and utilizes the reject sampling to derive a prioritized set of rubric items ("must-ask" items) for each case. We perform a comprehensive evaluation of state-of-the-art models and demonstrate that, across multiple assessment dimensions, current models face substantial challenges. Our results indicate that improving medical dialogue will require advances in dialogue management architectures, not just incremental tuning of the base-model.
\end{abstract}

\section{Introduction}
\begin{figure}
    \centering
    \includegraphics[width=\linewidth]{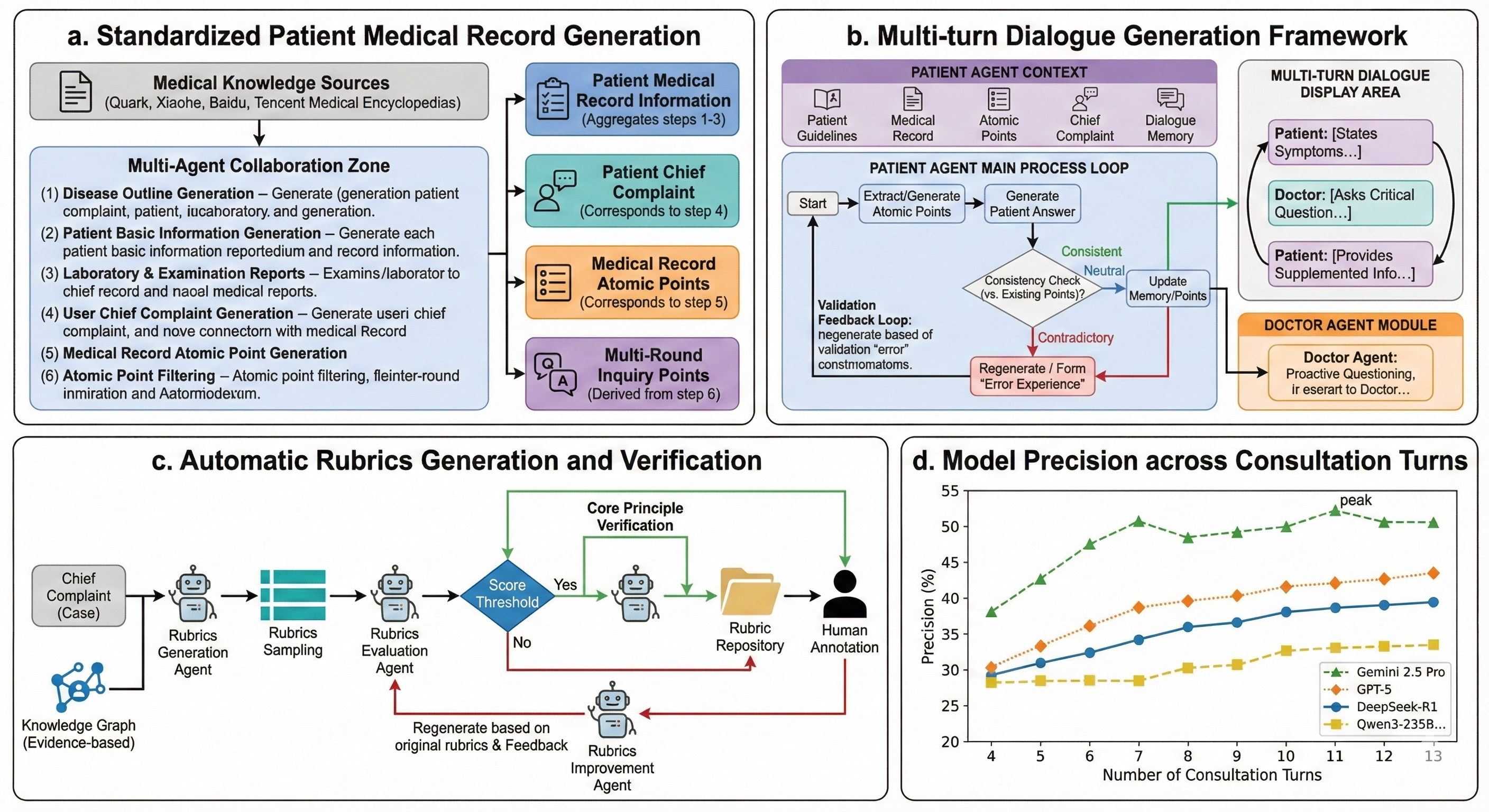}
    \caption{The landscape of MedDialogRubrics framework}
    \label{fig:landscape}
\end{figure}
Large Language Models (LLMs) have demonstrated remarkable capabilities in natural language understanding and generation, catalyzing a paradigm shift in medical informatics. Applications ranging from clinical decision support to patient-facing assistants promise to democratize healthcare access and alleviate clinician burnout \cite{jiang2023llmhealthcare,singhal2023clinicalknowledge}. However, safe deployment of LLMs in clinical settings requires evaluation frameworks that rigorously mirror the complexity of real-world medical practice.

In authentic clinical environments, diagnostic reasoning is not a static question-answering task but a dynamic, multi-turn process. Clinicians must navigate an open problem space, actively ask for information, formulate hypotheses, and refine their differential diagnosis through iterative interaction \cite{kassirer1991learning, nezhad2023diagnosing}. Despite this, existing benchmarks for medical LLMs focus predominantly on static tasks—such as multiple-choice questions (e.g. MedQA, MedMCQA) or summarization \cite{jin2020whatdisease,jin2022clinicallongformer}. Although these benchmarks evaluate domain knowledge retention, they fail to assess an LLM's ability to conduct a structured consultation, manage dialogue flow, or exhibit safety behaviors during information gathering.

Developing a robust evaluation framework for multi-turn medical consultations presents three distinct challenges:
\begin{itemize}
    \item \textbf{Data Scarcity and Privacy:} High-quality, multi-turn medical dialogue data is scarce due to strict privacy regulations (e.g., HIPAA/GDPR), limiting the scale of reproducibility studies.
    \item \textbf{Simulation reliability:} While agent-based simulation offers an alternative that preserves privacy, standard LLM-based patient agents are prone to hallucinations—inventing symptoms or contradicting medical logic—which compromising the validity of the evaluation \cite{nori2023gpt4medical}.
    \item \textbf{Evaluation Subjectivity:} Unlike multiple-choice questions with a single gold standard, evaluating a dialogue trajectory requires assessing whether "must-ask" questions were posed at the right time. This typically relies on costly human annotation or vague heuristic-based metrics.
\end{itemize}

To address these limitations, we introduce \textbf{MedDialogRubrics}, a comprehensive benchmark and evaluation framework designed to rigorously assess the diagnostic reasoning and information-gathering capabilities of LLMs. Unlike prior works that rely on unstructured simulations, our framework employs a clinically grounded, multi-agent synthesis pipeline. We construct a robust \textit{Patient Agent} anchored to atomic medical facts and augmented with a dynamic guidance mechanism to detect and correct hallucinations in real-time. Furthermore, we automate the evaluation process by generating over 60,000 fine-grained rubrics derived from Evidence-Based Medicine (EBM) guidelines, utilizing reject sampling to ensure clinical relevance.

Our key contributions are summarized as follows:
\begin{itemize}
    \item \textbf{A Novel Benchmark Dataset:} We release a dataset of 5,200 synthetically constructed, clinically coherent patient cases covering a diverse range of diseases, synthesized without accessing private real-world records.
    \item \textbf{Hallucination-Free Patient Simulation:} We propose a controlled Patient Agent architecture that decouples medical knowledge from dialogue generation, ensuring internal consistency and clinical plausibility during multi-turn interactions.
    \item \textbf{EBM-Grounded Evaluation Pipeline:} We introduce a structured rubric-generation pipeline that combines LLM retrieval with expert refinement, establishing a "must-ask" criterion for objective scoring.
    \item \textbf{Comprehensive Analysis:} We evaluate state-of-the-art LLMs using our framework, revealing significant gaps in current dialogue management architectures and highlighting the necessity for systems that go beyond incremental instruction tuning.
\end{itemize}

\section{Related Work}

\subsection{Medical QA and Clinical Reasoning Benchmarks}
The evaluation of Large Language Models (LLMs) in the medical domain has evolved from static knowledge retrieval to complex, multi-turn clinical reasoning and agentic interactions.\cite{schmidgall_agentclinic_2025,li_mediq_2024,arora_healthbench_nodate} Early benchmarks, such as MedMCQA\cite{pal_medmcqa_2022}, established a foundation by providing over 194,000 high-quality multiple-choice questions (MCQs) from Indian medical entrance exams, focusing on a model's ability to reason across diverse healthcare topics in a single-turn format. Similarly, LLM-MedQA\cite{yang_llm-medqa_2025} leverages the established MedQA dataset to enhance performance through multi-agent architectures and case study generation, emphasizing domain-specific terminology and zero-shot reasoning.

As the field progressed toward assessing clinical utility, Med-PaLM 2\cite{singhal_towards_2023} introduced a rigorous framework for long-form question answering, utilizing nine physician-validated evaluation axes and adversarial datasets to probe the limits of model safety and accuracy. To address the limitations of static QA, several benchmarks transitioned to multi-turn dialogue systems. LLM-Mini-CEX\cite{shi_llm-mini-cex_2023} adapted the traditional clinical exercise into an automated evaluation of diagnostic conversations, employing patient simulators to test both diagnostic capabilities and humanistic qualities. Liao et al. further advanced this by reformulating USMLE questions into interactive consultations, requiring models to proactively elicit missing patient information\cite{liao_automatic_2023,Liao_Meng_Wang_Liu_Wang_Wang_2024}. MediQ\cite{li_mediq_2024} specifically targets this proactive information-seeking behavior, simulating interactions between a Patient System and an Expert System to ensure reliable clinical reasoning under incomplete context.
Recent developments have introduced agentic and large-scale physician-led benchmarks. AgentClinic presents a multimodal agent environment that treats clinical decision-making as a sequential task involving external tools and electronic health records, revealing that interactive diagnosis is significantly more challenging than static answering. HealthBench\cite{arora_healthbench_nodate} represents one of the most comprehensive efforts to date, featuring 5,000 multi-turn conversations evaluated against over 48,000 unique rubric criteria validated by 262 physicians. Finally, MAQuE\cite{gong_dialogue_2025} focuses on the nuances of the "doctor agent," evaluating inquiry proficiency and patient experience across 3,000 simulated patient agents with diverse emotional and linguistic patterns. Together, these benchmarks reflect a paradigm shift toward ensuring LLMs can safely and effectively navigate the dynamic nature of real-world clinical practice. If evaluating a doctor were like a driving test, MedMCQA would be the written theory exam, while AgentClinic, MediQ, and HealthBench would be the actual road test, requiring the driver to handle unexpected traffic, talk to passengers, and navigate a changing environment safely.

\begin{table*}[htbp]
\centering
\caption{Comparison of existing medical consultation benchmarks.} 
\label{tab:benchmark_comparison} 
\renewcommand{\arraystretch}{1.2} 
\begin{tabular}{lcccccc}
\toprule
\textbf{Benchmark} & \textbf{\makecell{Multi-turn \\ Support}} & \textbf{\makecell{Key Points \\  \/Rubrics}} &  \textbf{Expert Validated} & \textbf{\makecell{Number of \\ \# Rubrics}} \\
\midrule
MedMCQA         & \xmark & \xmark & \xmark & -- \\
Med-PaLM 2      & \cmark & \xmark & \cmark & -- \\
LLM-Mini-CEX    & \cmark & \cmark & \cmark & -- \\
Liao et al.     & \cmark & \xmark & \xmark & -- \\
AgentClinic     & \cmark & \xmark & \xmark & -- \\
MediQ           & \cmark & \xmark & \xmark & -- \\
LLM-MedQA       & \xmark & \xmark & \xmark & -- \\
HealthBench     & \cmark & \cmark & \cmark & 48,562 \\
MAQuE           & \cmark & \xmark & \xmark & -- \\
\midrule
\textsc{MedDialogRubrics} (Ours) & \cmark & \cmark & \cmark & 60,000 \\
\bottomrule
\end{tabular}
\end{table*}

Some recent efforts attempt to incorporate more dynamic reasoning tasks. For example, interactive patient simulators have been used in medical education for trainee evaluation \cite{cook2011simulatedpatients}, and some LLM studies explore chain-of-thought or stepwise diagnostic reasoning \cite{nezhad2023diagnosing,nori2023gpt4medical}. However, existing simulators are either proprietary, manually constructed, or not designed for large‑scale LLM evaluation.

\subsection{Dialogue Simulation and Patient Agents}
Dialogue simulators in general NLP research include user simulators for task‑oriented dialogue in domains such as travel booking or customer service \cite{schatzmann2006surveyusersimulation,budzianowski2019gpt3dialog}. However, medical contexts require stricter factual consistency, symptom logic, and safety considerations. Some recent work proposes LLM‑driven patient agents \cite{wang2023patientsimulator}, but these often lack controls for reproducibility or clinical correctness, making them unsuitable for standardized evaluation in all models.

Recent advancements in Large Language Models (LLM) have shifted the focus of medical AI from static question-answering \cite{fan-etal-2025-ai,sviridov20253mdbench} to autonomous agentic systems capable of simulating complex, multi-turn clinical interactions.\cite{info16100894,zhi_reinventing_2025} Central to this shift is the development of "patient agents"—LLM-based entities designed to replicate the biological, psychological, and communicative traits of human patients.\cite{kyung2025patientsimpersonadrivensimulatorrealistic,wang2024patient} While early work relied on simple persona-based prompting \cite{cook2025creating}, current research has pivoted toward more sophisticated construction paradigms to enhance behavioral realism and factual grounding.

A prominent methodological trend is the use of persona-driven frameworks like PatientSim, which models behavior along axes of personality, literacy, and cognitive confusion.\cite{kyung2025patientsimpersonadrivensimulatorrealistic} To ensure clinical validity, these models are increasingly aligned using Direct Preference Optimization (DPO), which significantly outperforms Supervised Fine-Tuning (SFT) in handling complex reasoning and emotional nuance\cite{liu2025eeyore}. Furthermore, the integration of Retrieval-Augmented Generation (RAG) and Multi-Agent Systems (MAS) has enabled agents to interact with simulated Electronic Health Records (EHR) and external diagnostic tools.\cite{info16100894, ai6090226, fan-etal-2025-ai} Frameworks such as AI Hospital \cite{fan-etal-2025-ai} and MedAgentSim \cite{almansoori2025selfevolvingmultiagentsimulationsrealistic} have established multi-role environments (Doctor, Patient, Examiner) to evaluate diagnostic accuracy within realistic clinical workflows.

The evaluation of these agents has similarly evolved from linguistic metrics (BLEU/ROUGE) to action-oriented benchmarks like MedAgentBench \cite{jiang2025medagentbenchrealisticvirtualehr} and MedAgentBoard.\cite{zhu2025medagentboardbenchmarkingmultiagentcollaboration} These benchmarks challenge agents to execute multi-step clinical tasks in FHIR-compliant environments, highlighting a "reliability gap" in current state-of-the-art models for action-based vs. query-based tasks.Emerging research now targets the "statelessness" of current agents, proposing Bayesian-inspired memory management systems to maintain longitudinal patient models for chronic disease simulation\cite{zhi_reinventing_2025,lu2025dynamicaffectivememorymanagement}. Collectively, these developments lay the groundwork for high-fidelity, autonomous patient simulators that bridge the gap between AI-driven medical knowledge and its practical application in dynamic clinical settings.

\begin{table*}[htbp]
\centering
\caption{Comparison of Dialogue Simulation and Patient Agents Framework} 
\label{tab:simulation_comparison} 
\renewcommand{\arraystretch}{1.2} 
\begin{tabular}{lcccc}
\toprule
\textbf{Method} & \textbf{\makecell{Implementation \\ Cost}} & \textbf{\makecell{Medical \\ Accuracy}} &  \textbf{\makecell{Behavioral \\ Realism}} & \textbf{\makecell{Consistency}} \\
\midrule
Prompt Engineering & Very Low  & Moderate  & High      & Low \\
RAG                & Moderate  & Very High & Moderate  & High \\
Fine-Tuning        & High      & High      & Very High & Moderate \\
Multi-Agent        & Very High & High      & Moderate  & Very High \\
\midrule
\end{tabular}
\end{table*}

\subsection{Automated Evaluation for Medical LLMs (LLM-as-a-Judge)}
 While traditional n-gram metrics (e.g. ROUGE, BLEU) remain insufficient for capturing the clinical validity of generated text, the "LLM-as-a-Judge" paradigm has emerged as a scalable alternative to costly human expert annotation. Recent benchmarks such as HealthBench\cite{arora_healthbench_nodate} and MedHELM\cite{bedi2025medhelm} have shifted the focus of evaluation from rote knowledge recall to rubric-based behavioral assessment and jury-based consensus, respectively. Specifically, HealthBench utilizes over 48,000 granular criteria to assess safety and adherence, whereas MedHELM employs an ensemble of judge models to mitigate individual model biases. Despite their promise, studies by Maina et al. \cite{williams2025human} highlight persistent challenges, including substantial verbosity bias, inconsistency in low-resource languages (e.g., Kinyarwanda), and a "severity gap" where models like GPT-5 and Gemini exhibit divergent leniency compared to human clinicians. Notably, reasoning-enhanced models such as DeepSeek-R1 and GPT-o3-mini have recently demonstrated superior inter-rater reliability with human experts compared to standard instruction-tuned models, suggesting that explicit reasoning chains are critical for reliable automated clinical evaluation.consultations with formal scoring criteria.

\subsection{Gaps in Current Research}
Across these areas, three major gaps emerge:
\begin{itemize}
    \item The lack of clinically grounded, scalable multi‑turn medical‑dialogue datasets with controlled patient behavior.
    \item The absence of structured benchmarks that evaluate the inquiry strategy and the progression of diagnostic reasoning.
    \item Limited rigor in applying LLM‑as‑judge methods to medical reasoning tasks, where errors may have significant safety implications.
\end{itemize}
Our framework directly addresses these gaps by combining synthetic case generation, structured clinical key‑point annotation, a reproducible patient agent, and a calibrated LLM‑as‑judge evaluation pipeline. To our knowledge, it is the first fully integrated benchmark specifically designed for evaluating multi‑turn medical consultation competence in LLMs.

\section{Methods}
\subsection{Phase 1: Multi-Agent Patient Record Generation}
To facilitate scalable and reproducible evaluation of multi-turn diagnostic interactions, we adopt the Patient-Zero framework~\cite{lai_patient-zero_2025}. This approach enables the generation of synthetic patient data without accessing real-world medical records, thereby strictly preserving patient privacy. The generation pipeline consists of three distinct phases:

\begin{itemize}
    \item \textbf{Disease Knowledge Retrieval:} We source comprehensive disease-specific knowledge—including symptoms, epidemiology, and complications—from authoritative open medical encyclopedias.
    
    \item \textbf{Multi-Step Record Generation:} A hierarchical multi-agent system constructs the patient record through three progressive steps:
    \begin{enumerate}
        \item \textit{Disease Outline}: A high-level summary of the clinical presentation.
        \item \textit{Basic Information}: Demographics, lifestyle factors, and medical history.
        \item \textit{Detailed Clinical Data}: Specifics regarding symptom duration, severity, and pertinent negative findings.
    \end{enumerate}
    
    \item \textbf{Chief Complaint Synthesis:} Based on the fully generated record, a specialized agent synthesizes the patient's "Chief Complaint" (the initial statement presented to the physician) to simulate realistic consultation initiation.
\end{itemize}

\begin{figure}
    \centering
    \includegraphics[width=1\linewidth]{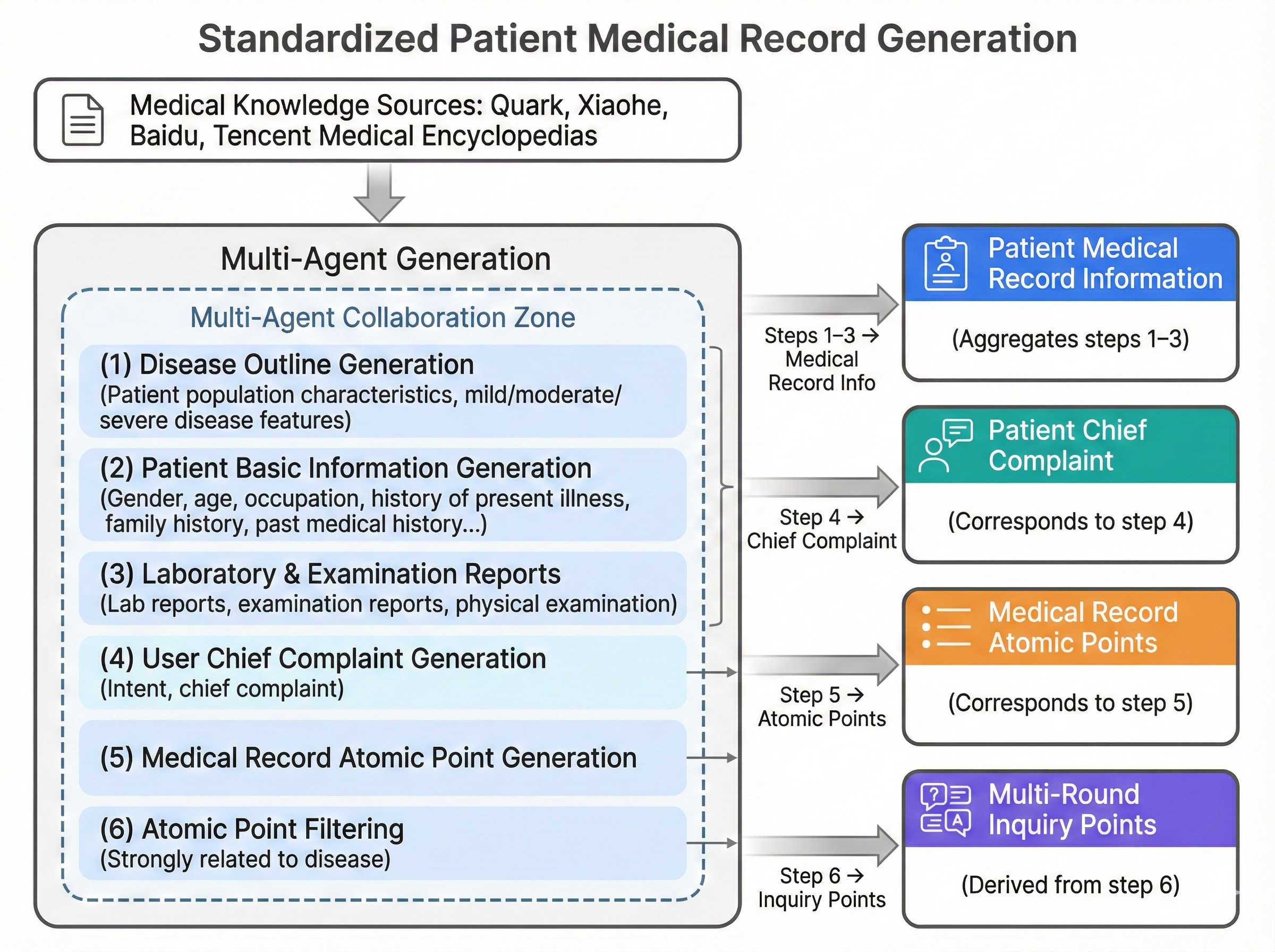}
    \caption{Multi-Agent Patient Record Generation Pipeline}
    \label{fig:record}
\end{figure}
\paragraph{Stage 1: Disease Knowledge Retrieval}
We begin by selecting the target conditions from a curated taxonomy that covers common primary care presentations, chronic diseases, mental‑health conditions, and acute emergencies. For each condition, we define:
\begin{itemize}
    \item Core symptoms (required for diagnosis)
    \item Auxiliary symptoms (optional but common)
    \item Red‑flag symptoms that signal severe or urgent presentations
    \item Risk factors, epidemiological constraints, and typical disease trajectory
    \item Differential diagnoses with overlapped symptoms
\end{itemize}

These elements are derived from clinical guidelines and medical knowledge sources (e.g., UpToDate, NICE guidelines, standard textbooks, Baidu Health Encyclopedia\footnote{\hyperlink{Baidu Health Encyclopedia}{https://expert.baidu.com/dict/pages/main/index}}, Kuake Health Encyclopedia). Each disease profile is encoded in a structured template.

\paragraph{Stage 2: Multi-Step Record Generation}

Once a disease profile is established, we sample distributions based on demographics, severity of symptoms, onset time, comorbidities, and lifestyle variables. Rules guaranty internal consistency, such as ensuring that the progression of symptoms aligns with the chosen stage of the disease or that incompatible comorbid conditions do not coexist.

A multi-agent system generates the record in stages:

\begin{itemize}
    \item Disease Overview: A high-level summary of the presentation.
    \item Basic Information: Demographic, lifestyle, and medical history.
    \item Detailed Clinical Data: duration, severity, and findings of negative symptoms.
\end{itemize}

\paragraph{Stage 3: Chief Complaint Synthesis}

Based on the complete record, a separate agent synthesizes the patient's initial "Chief Complaint" (the opening statement to the doctor).

Then We apply automated validators that check for:
\begin{itemize}
    \item Missing required symptoms
    \item Contradictory time courses
    \item Misaligned demographics and risk factors
    \item Overlap between the case description and distractor illnesses
\end{itemize}

Cases that fail these checks are revised or discarded. The result is a dataset of high‑quality diagnostic scenarios.

\subsection{Phase 2: Patient Agent Design}
\begin{figure}
    \centering
    \includegraphics[width=\linewidth]{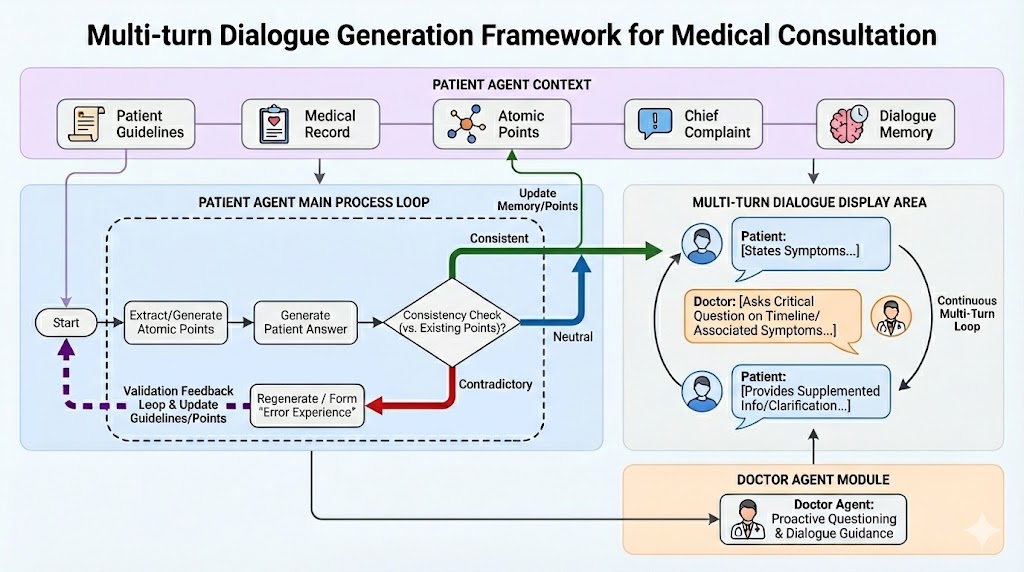}
    \caption{Patient Agent \& Multi-turn Dialogue Generation Framework}
    \label{fig:patient}
\end{figure}

A central component of the evaluation framework is the patient agent, which simulates patient behavior during multi-turn interactions. The agent must be realistic enough to challenge the model while remaining deterministic to provide a fair comparison between models.

As shown in \ref{fig:patient}, the core design features of the patient agent include:

\begin{itemize}
    \item The Patient Agent is designed to simulate a real patient interacting with the candidate Doctor LLM.
    \item Atomic Decomposition: The generated narrative record is decomposed into a set of discrete "atomic statements" (e.g., Symptom: Headaches, Duration: 2 weeks, Trigger: Bright light). This serves as the agent's "ground truth" memory.
    \item Strict Adherence and Inference: The agent is prompted to answer questions strictly based on these atomic facts. If asked about missing details, it provides plausible answers that do not contradict the record (e.g., denying a specific allergy if the record lists "No known allergies").
    \item Guidance Injection Loop: To prevent hallucinations observed in standard agents, we implement a feedback loop. If the Patient Agent generates a response that conflicts with its atomic memory during a turn, the system catches the error, suppresses the output, and injects a "Guidance Prompt" (e.g., Warning: Your record states you are a non-smoker. Correct your response.). This guidance persists in the context of preventing repeated errors.
\end{itemize}

\subsection{Phase 3: LLM-Based \& Expert-Annotation EBM-Driven Rubric Generation}
To systematically evaluate the multi-turn dialogue, we define a gold‑standard list of key inquiry rubrics for each case. These represent the essential information a competent clinician would seek during the diagnostic interview. The key inquiry rubrics can be seen in Table \ref{tab:rubric_criteria}.

To enhance factual grounding and controllability, the key rubrics are derived using a hybrid manual–automated methodology: large language models (LLMs) first generate candidate key points from the case description, after which clinical experts systematically review, filter, and annotate these candidates.

\subsubsection{Generation of Multi-Agent rubrics with rejection Sampling}
\begin{figure}
    \centering
    \includegraphics[width=1\linewidth]{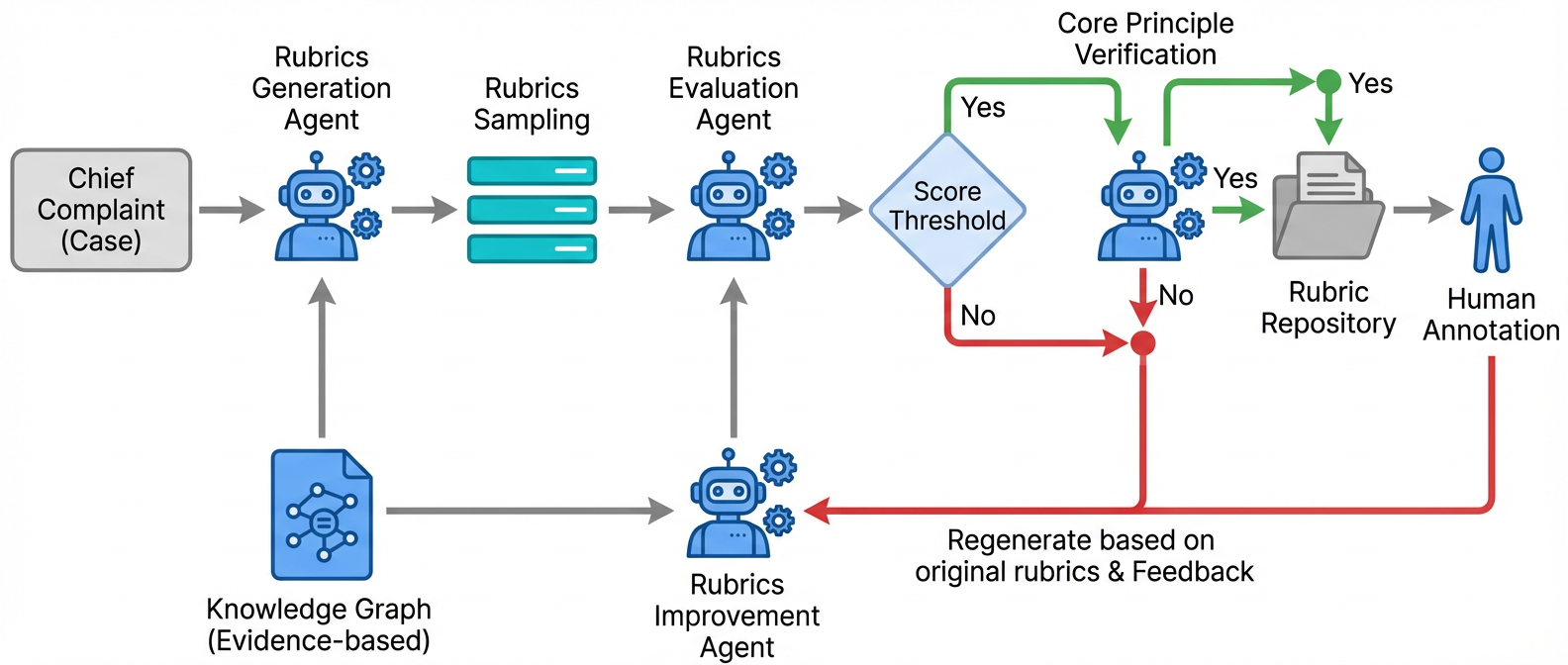}
    \caption{Generation of Multi-Agent rubrics with rejection Sampling}
    \label{fig:placeholder}
\end{figure}

As illustrated in Figure~\ref{fig:placeholder}, we propose a multi-agent pipeline with (i) a \textbf{Rubric Generation Agent} that proposes candidate rubrics grounded in an evidence-based knowledge graph, (ii) a \textbf{Rubric Evaluation Agent} that scores candidates with respect to predefined quality criteria, (iii) \textbf{reject sampling} with thresholding to filter low-quality candidates, and (iv) a \textbf{Rubric Improvement Agent} that iteratively refines rejected candidates using structured feedback. Accepted and verified rubrics are stored in a \textbf{rubric repository} for downstream \textbf{human annotation}.

\paragraph{Evidence Grounding via Knowledge Graph.}
We assume access to an evidence-based knowledge graph $\mathcal{G}$ (e.g., guideline-derived diagnostic pathways) that encodes clinically relevant entities and relations. For an input case $x$, we retrieve a subgraph $\mathcal{G}_x$ (or a set of relevant triples) to constrain generation and promote coverage of guideline-mandated inquiry dimensions.

\paragraph{Rubric Generation Agent.}
The generation agent parameterized by an LLM produces a candidate rubric set
\begin{equation}
\mathcal{R}^{(k)} \sim p_{\theta}\!\left(\mathcal{R}\mid x, \mathcal{G}_x\right),
\end{equation}
where we sample $K$ candidates $\{\mathcal{R}^{(k)}\}_{k=1}^{K}$ to encourage diversity and avoid mode collapse. In practice, $\mathcal{R}^{(k)}$ may be generated via temperature-controlled sampling and/or prompting constraints derived from $\mathcal{G}_x$.

\paragraph{Rubric Evaluation Agent.}
An independent evaluator agent assigns each candidate set $\mathcal{R}^{(k)}$ a scalar score $s^{(k)}\in\mathbb{R}$:
\begin{equation}
s^{(k)} = f_{\phi}\!\left(x, \mathcal{G}_x, \mathcal{R}^{(k)}\right),
\end{equation}
where the scoring rubric aggregates criteria commonly required in clinical interviewing, including: (i) relevance to $x$, (ii) coverage/completeness, (iii) non-redundancy, (iv) clarity/actionability, and (v) consistency with evidence in $\mathcal{G}_x$.

\paragraph{Core Principle Verification.}
Beyond scalar scoring, we enforce a set of hard constraints (``core principles'') via a verifier agent $v(\cdot)$:
\begin{equation}
\textsc{Verify}\!\left(\mathcal{R}^{(k)}\right) \in \{0,1\},
\end{equation}
which checks, for example, that guideline-mandated red flags are covered, unsafe/irrelevant inquiries are absent, and each rubric is grounded (directly or indirectly) in $\mathcal{G}_x$.

\paragraph{Rubric Improvement Agent.}
For candidates that are rejected or fail verification, an improvement agent revises rubrics conditioned on structured feedback:
\begin{equation}
\widetilde{\mathcal{R}}^{(k)} \sim p_{\psi}\!\left(\mathcal{R}\mid x,\mathcal{G}_x,\mathcal{R}^{(k)}, \mathcal{F}^{(k)}\right),
\end{equation}
where $\mathcal{F}^{(k)}$ contains evaluator/verifier feedback (e.g., missing red flags, redundancy, ambiguous phrasing). The refined rubrics are re-submitted to evaluation and verification, forming a closed-loop refinement process.

\paragraph{Reject Sampling.}
We apply reject sampling by accepting a candidate if its evaluation score exceeds a threshold $\tau$:
\begin{equation}
\textsc{Accept}\!\left(\mathcal{R}^{(k)}\right) \triangleq \mathbb{I}\left[s^{(k)}\ge \tau\right].
\end{equation}
A candidate is stored only if it is both accepted and passes core principle verification:

\begin{equation*}
\begin{aligned}
\mathcal{R}^{(k)} \in \mathcal{D}
\iff\;& \textsc{Accept}\!\bigl(\mathcal{R}^{(k)}\bigr)=1 \wedge\ \textsc{Verify}\!\bigl(\mathcal{R}^{(k)}\bigr)=1 .
\end{aligned}
\end{equation*}

where $\mathcal{D}$ denotes the rubric repository.

\begin{algorithm}[t]
\caption{Multi-Agent Rubric Generation with Reject Sampling}
\label{alg:rubric}
\begin{algorithmic}[1]
\REQUIRE Case description $x$; knowledge graph $\mathcal{G}$; retrieval function $\textsc{Retrieve}$; generator $p_{\theta}$; evaluator $f_{\phi}$; verifier $v$; improver $p_{\psi}$; threshold $\tau$; samples $K$; max refinement steps $T$.
\ENSURE Verified rubric set(s) added to repository $\mathcal{D}$.
\STATE $\mathcal{G}_x \leftarrow \textsc{Retrieve}(\mathcal{G}, x)$
\FOR{$k=1$ \TO $K$}
    \STATE $\mathcal{R}^{(k)} \sim p_{\theta}(\mathcal{R}\mid x,\mathcal{G}_x)$
    \FOR{$t=1$ \TO $T$}
        \STATE $s^{(k)} \leftarrow f_{\phi}(x,\mathcal{G}_x,\mathcal{R}^{(k)})$
        \STATE $a \leftarrow \mathbb{I}[s^{(k)} \ge \tau]$
        \STATE $b \leftarrow v(x,\mathcal{G}_x,\mathcal{R}^{(k)})$
        \IF{$a=1$ \AND $b=1$}
            \STATE $\mathcal{D} \leftarrow \mathcal{D} \cup \{\mathcal{R}^{(k)}\}$ \COMMENT{Store verified rubrics}
            \STATE \textbf{break}
        \ELSE
            \STATE $\mathcal{F}^{(k)} \leftarrow \textsc{Feedback}(s^{(k)}, a, b)$
            \STATE $\mathcal{R}^{(k)} \sim p_{\psi}(\mathcal{R}\mid x,\mathcal{G}_x,\mathcal{R}^{(k)},\mathcal{F}^{(k)})$
        \ENDIF
    \ENDFOR
\ENDFOR
\RETURN $\mathcal{D}$
\end{algorithmic}
\end{algorithm}

\subsubsection{Expert Annotation}

\begin{figure}
    \centering
    \includegraphics[width=1\linewidth]{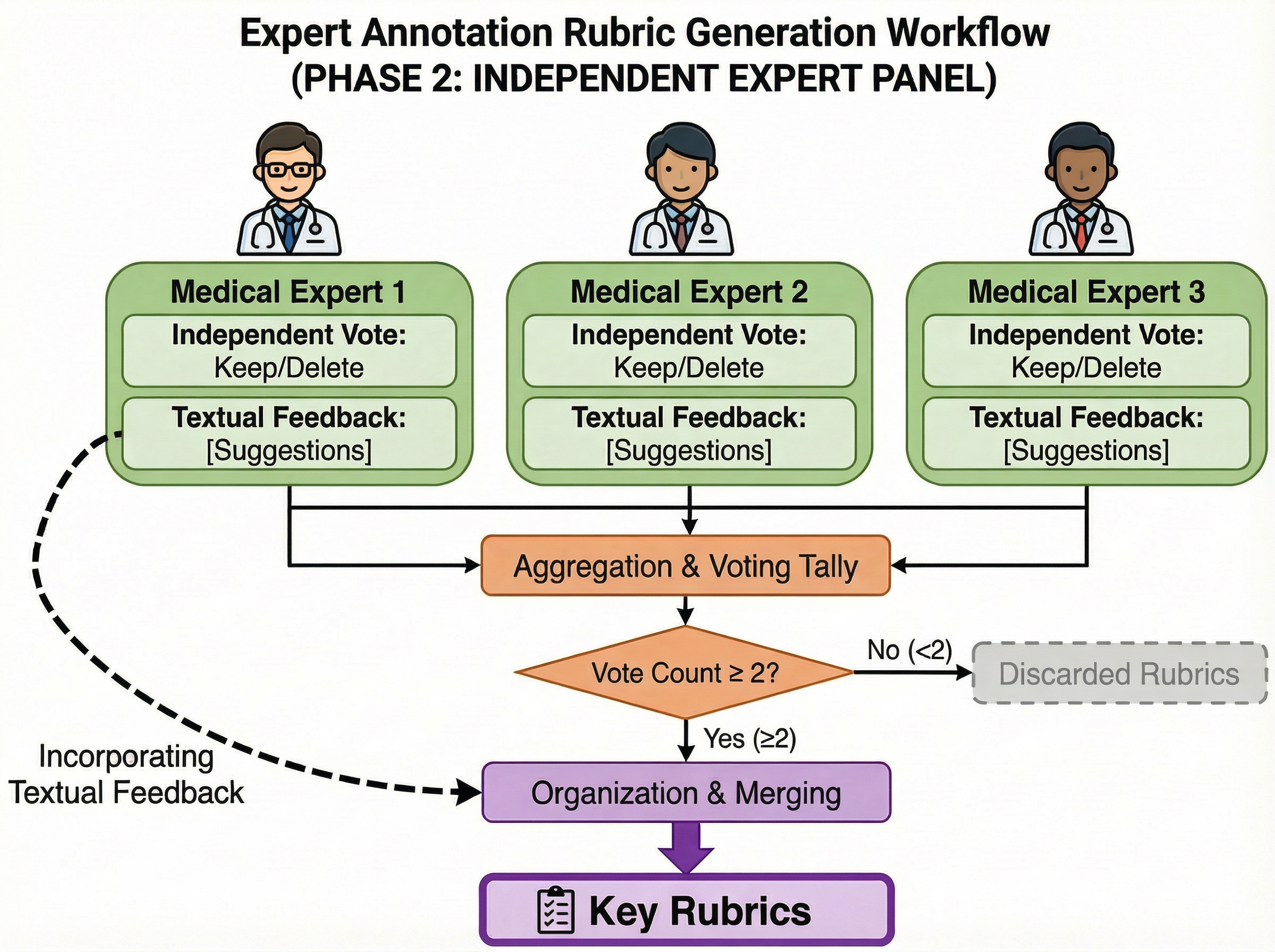}
    \caption{Expert Annotation Rubric Generation Workflow}
    \label{fig:expertannotation}
\end{figure}

As shown in Figure \ref{fig:expertannotation}, an independent panel of three domain experts evaluates each candidate rubric in parallel. For every rubric, each expert provides (i) an independent binary vote (\emph{Keep} vs.\ \emph{Delete}) and (ii) free-form textual feedback containing concrete revision suggestions (e.g., clarifications, scope constraints, wording edits, or edge-case handling).

All votes are then aggregated and tallied. A rubric is retained for further processing only if it receives at least two \emph{Keep} votes (majority agreement; vote count $\geq 2$). Rubrics failing to reach this threshold ($<2$ \emph{Keep} votes) are discarded.

For retained rubrics, the system performs organization and merging: overlapping or semantically redundant rubrics are consolidated, and the retained rubrics are refined by incorporating the experts' textual feedback. This consolidation-and-revision stage yields a curated set of \emph{Key Rubrics}, representing the final expert-validated rubric set produced in this phase.

\subsection{Phase 4: LLM-as-Judge Evaluation Framework}

We developed a structured LLM-as-judge evaluation framework designed to provide clinically grounded, reproducible, and bias-minimized assessments of multi-turn diagnostic dialogues. The framework was informed by empirical performance patterns observed in three independent datasets (7166 matched samples), ensuring that the adjudication faithfully reflects real-world behaviors and safety requirements in medical reasoning.

Each conversation from the interactions between different doctor agents and simulated patient agents is evaluated against all the rubric criteria for each case using a model as a grader, in an LLM-as-ajudge setup. The model-based grader outputs binary judgment verdicts for each rubric, which are {Satisfied and Not Satisfied}. The final case score is the weighted sum of all positive and negative weights, normalized by the sum of the positive weights (the maximum possible score that the doctor agent can achieve).

\begin{equation}
S_k = \frac{\sum_{r_i\in R} I_{r_i} m_{r_i}}{\sum_{r_i\in R} I_{r_i}},
m_{r_i}=\mathrm{Judge}(C_k, r_i)
\end{equation}

\begin{equation}
  m_{r_i} =
  \begin{cases}
    1, & \text{if } r_i \text{ is satisfied},\\
    0, & \text{if } r_i \text{ is not satisfied}.
  \end{cases}
\end{equation}

where $S_k$ is the final case score for the case and the conversation $C$ generated by the doctor agents. R is the set of all rubrics for each case $k$, $I_{r_i}$ is the importance score assigned to rubric $r_i$ and $m_{r_i}$ is the binary indicator returned from the model-based judge, $\mathrm{Judge}(\cdot, \cdot)$, representing the llm-as-judge process for rubric $r_i$.

\paragraph{Human Consistency Analysis}
Similar to HealthBench~\cite{arora_healthbench_nodate}, we utilize the Macro F1 score to validate the effectiveness of using a model-based grader as a proxy for human judgment. In our setup, we compare the ground truth judgment of experts and model-based graders for each task and compute the F1 scores for each of the classes \{{Satisfied}, {Not Satisfied\}}.

\begin{equation}
\begin{aligned}
F_{1} &= 2 \cdot \frac{\mathrm{precision}\cdot \mathrm{recall}}
{\mathrm{precision}+\mathrm{recall}},\\
\mathrm{precision} &= \frac{TP}{TP+FP},
\mathrm{recall} = \frac{TP}{TP+FN},
\end{aligned}
\label{eq:f1}
\end{equation}
where $TP$, $FP$, and $FN$ are the True Positive, False Positive, and False Negative values, respectively. We also performed ablation studies to isolate the most significant factors in the level of alignment between the model-based grader and human judgments. For more details, see Section~\ref{reliability}.

\section{Experiments Results and Analysis}
We evaluate the medical consultation capabilities of four representative Large Language Models (LLMs)—spanning both proprietary and open-source paradigms—functioning as doctor agents within the MedDialogRubrics framework. Distinguishing itself from existing benchmarks, our framework incorporates over 60,000 expert-annotated rubric criteria across more than 4,700 cases. This high-resolution granularity facilitates an atomic-level quality assessment, enabling the precise identification of specific failure modes that remain obscured by traditional, coarse-grained evaluation metrics.

\subsection{Experimental Setup}

\paragraph{Evaluated Systems} We benchmark two state-of-the-art open-source models (Qwen3-235B-A22B-Instruct-2507\cite{qwen2.5-1m}, DeepSeek-R1\cite{deepseekai2025deepseekr1incentivizingreasoningcapability}) and two state-of-the-art proprietary models (GPT-5\cite{openai_gpt5_2025}, Gemini-2.5-Pro\cite{comanici2025gemini}). 

This study focuses specifically on assessing the inquiry completeness of large language models (LLMs) in a clinical context. To this end, we configure each model to conduct multi-turn clinical interviews with a simulated patient agent that generates responses conditioned on an underlying electronic medical record. Using a unified system prompt that explicitly specifies the evaluation dimensions, we instruct each model to adopt the role of a physician and perform a multi-round clinical history-taking process. At each dialogue turn, the model receives the full interaction history and may either continue asking questions or terminate the consultation by outputting the token “End Inquiry.” To avoid unbounded interactions, we cap each session at a maximum of 12 turns. 

For both patient simulation and inquiry quality evaluation, we employ DeepSeek-V3, selected based on its favorable trade-off between performance and computational cost. The detailed prompting schemes used for all components are described below.

\paragraph{Ensemble Strategies and Implementation}
Evaluating models in isolation fails to capture the complexity of medical decision-making. Therefore, we implement robust ensemble strategies using a panel of three advanced LLMs: GPT-5, Gemini-2.5-Pro, and DeepSeek-V3\cite{deepseekai2024deepseekv3technicalreport}. The alignment of these judges is rigorously verified against human experts (via Macro F1 on 300 cases). To simulate clinical decision boards, we assess three distinct aggregation mechanisms:
\begin{itemize}[noitemsep, topsep=0pt]
    \item \textbf{Majority Voting:} Reflects democratic consensus to filter outliers.
    \item \textbf{Unanimous Voting:} Enforces a zero-tolerance policy for disagreement to ensure high precision.
    \item \textbf{Liberal Aggregation:} Prioritizes recall by flagging a condition if any single agent detects it, mimicking safety-critical screening protocols.
\end{itemize}
These strategies allow us to quantify the reliability gains of deploying multi-LLM systems in diagnostic scenarios.

\paragraph{Evaluation Pipeline}
The evaluation of each doctor agent is conducted through the following procedure:
\begin{enumerate}
    \item A multi-turn consultation is first generated by interacting with the model and a controlled patient agent.
    \item Inquiry actions and reasoning patterns are then extracted from the resulting dialogue context.
    \item Structured scoring is applied utilizing the MedDialogRubrics LLM-as-a-Judge pipeline.
    \item Consistency verification is performed, with safety penalties enforced where discrepancies arise.
    \item Scores are aggregated at both the per-case and per-dataset granularities.
    \item Finally, classical metrics—including Precision, Recall, Accuracy, and F1-score—are computed to quantify performance.
\end{enumerate}

\begin{figure}
    \centering
    \includegraphics[width=1\linewidth]{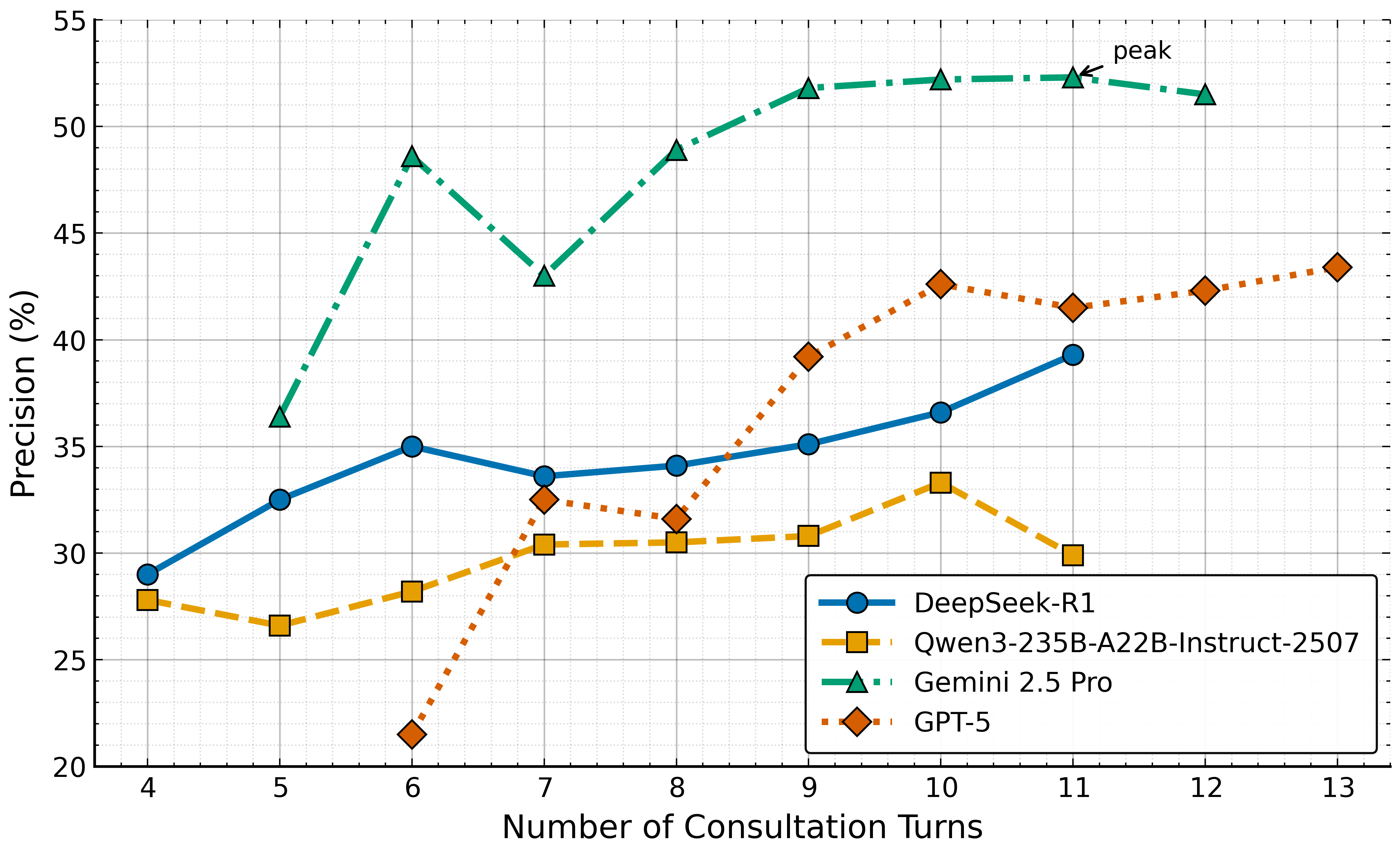}
    \caption{Relationship between the number of consultation turns and rubric match precision for different LLM-based doctor–patient agent pairs. Each curve reports the precision of matching asked consultation questions to the reference consultation key points prior to reaching a final conclusion, illustrating how information coverage and alignment evolve as dialogue length increases across models.}
    \label{fig:comparison}
\end{figure}

\subsection{Analysis of Information Gathering and Temporal Dynamics}
Figure \ref{fig:comparison} illustrates the trajectory of diagnostic accuracy relative to the length of the consultation (number of turns). We observe distinct behavioral patterns across the evaluated models:
\paragraph{Efficiency vs. Exhaustiveness:} Gemini-2.5-pro (green trajectory) demonstrates superior information-seeking efficiency. It achieves a rapid increase in accuracy, peaking at approximately 52\% around turns 9-10 before stabilizing. This suggests a highly strategic inquiry policy that prioritizes high-value medical facts early in the conversation.
\paragraph{The "Late-Bloomer" Phenomenon:} In contrast, GPT-5 (red trajectory) exhibits a linear growth pattern. While it starts with a lower match rate compared to Gemini and DeepSeek at turn 4 ($<30\%$), it surpasses DeepSeek by turn 9 and continues to improve, achieving its highest performance at 13+ turns. This behavior indicates a more cautious or exhaustive reasoning strategy, requiring deeper context to formulate precise inquiries.
\paragraph{Performance Plateaus:} DeepSeek-R1 and Qwen3-235B-A22B-Instruct-2507 show more modest gains. DeepSeek-R1 shows steady but slow improvement, plateauing near 40\%, while Qwen3-235B-A22B-Instruct-2507 struggles to exceed a 30-35\% match rate regardless of dialogue length.
\paragraph{Key Takeaway:} The disparity between the strongest model (Gemini-2.5-pro at $\sim52\%$) and the theoretical maximum (100\%) highlights the "substantial challenges" mentioned in our abstract. Even advanced models miss nearly half of the critical diagnostic criteria defined by experts, underscoring the difficulty of the \textbf{MedDialogRubrics} benchmark.

\subsection{Ablation Study on Patient Agent Components}
To validate the effectiveness of the proposed Patient Agent framework, we conducted an ablation study assessing the impact of each component on agent fidelity. Table~\ref{tab:patient_behavior} reports the performance metrics across three progressive configurations: Basic, Strict Adherence, and the full pipeline with Guidance Injection.The \textbf{Basic} setup, which relies solely on prompt engineering without constraints, exhibits a relatively high hallucination rate ($0.129$) and suboptimal consistency in behavior ($0.689$). This suggests that a standard LLM struggles to maintain a consistent patient persona over multi-turn dialogues.Incorporating \textbf{Strict Adherence \& Inference} significantly enhances the agent's stability. By grounding responses in decomposed atomic statements, the \textit{Behavior} score improves dramatically from $0.689$ to $1.000$, and the hallucination rate decreases to $0.076$. This indicates that constraining the agent to "ground truth" memory is essential for simulating consistent patient logic.Finally, the deployment of the \textbf{Guidance Injection Loop} achieves the best performance. This feedback mechanism further attenuates the hallucination rate to $0.049$ and boosts \textit{Relevance} to $0.992$. Notably, under this full configuration, the agent achieves perfect scores ($1.000$) across all Anthropomorphism dimensions (Linguistics, Cognition, and Behavior). These results empirically demonstrate that our dual-mechanism design (Strict Adherence + Guidance Loop) effectively mitigates the stochastic hallucinations typical of LLMs while preserving high-fidelity patient simulation.

\begin{table*}[t]
\centering
\setlength{\tabcolsep}{2.5pt}
\renewcommand{\arraystretch}{1.15}
\begin{tabular}{lccccc}
\toprule
\multirow{2}{*}{\textbf{Patient Agent Component}} 
& \multirow{2}{*}{\textbf{Hallucination $\downarrow$}}
& \multirow{2}{*}{\textbf{Relevance $\uparrow$}}
& \multicolumn{3}{c}{\textbf{Anthropomorphism}} \\
\cmidrule(lr){4-6}
& & & \textbf{Linguistics $\uparrow$} & \textbf{Cognition $\uparrow$} & \textbf{Behavior $\uparrow$} \\
\midrule
Basic                    & 0.129 & 0.92 & 0.981 & 0.971 & 0.689 \\
\quad + Strict Adherence \& Inference     & 0.076 & 0.951 & 0.993 & 1.000 & 1.000 \\
\quad\quad + Guidance Injection Loop   & 0.049 & 0.992 & 1.000 & 1.000 & 1.000 \\
\bottomrule
\end{tabular}
\caption{Ablation study of the Patient Agent. We report the Hallucination rate ($\downarrow$), Relevance ($\uparrow$), and Anthropomorphism scores across three progressive configurations. The results demonstrate that the combination of Strict Adherence and the Guidance Injection Loop significantly minimizes hallucinations while maximizing behavioral realism.}
\label{tab:patient_behavior}
\end{table*}

\subsection{Human-LLM Judge Alignment for Auto-Evaluation}
\subsubsection{Ensemble strategies}
\begin{figure}
    \centering
    \includegraphics[width=\linewidth]{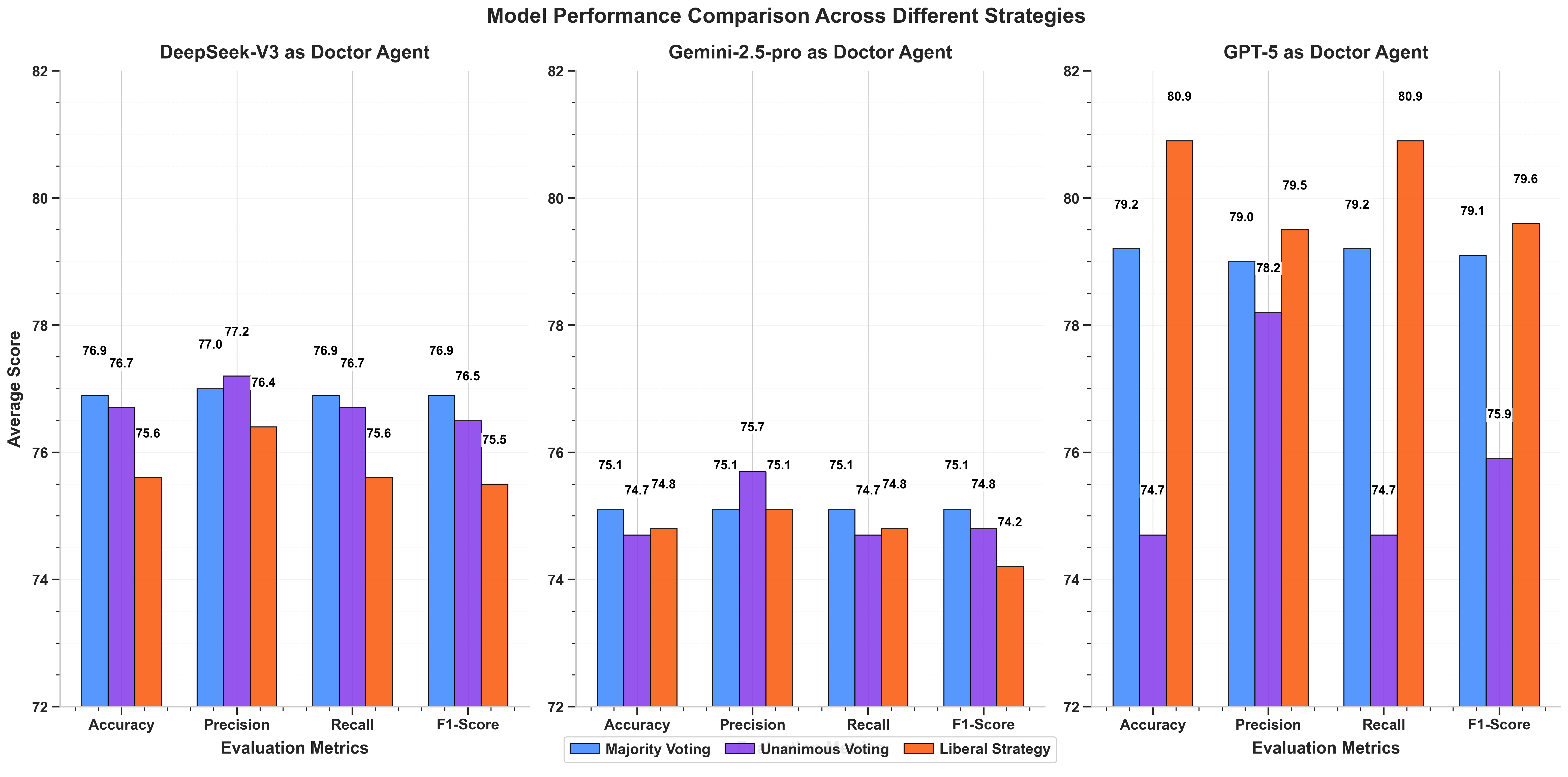}
    \caption{Evaluation metics on different doctor models using different judge strategies}
    \label{fig:autoevaluation}
\end{figure}

To ensure the scalability of our benchmark, we validated our automated scoring mechanism. Figure \ref{fig:autoevaluation} compares the alignment of three voting strategies—Majority Voting, Unanimous Voting, and Liberal Strategy—against ground-truth scores provided by clinical experts.
\paragraph{Robustness of Majority Voting:} Across all three Doctor Agents (DeepSeek, Gemini, GPT-5), the Majority Voting strategy (blue bars) consistently produces stable agreement with human experts, maintaining F1-scores in the 75–79\% range. This confirms that an ensemble of LLM judges can reliably approximate human clinical judgment.
\paragraph{Precision-Recall Trade-offs:} The Unanimous Voting strategy (purple bars) typically results in lower Recall and F1-scores (e.g., dropping to $\sim74.7\%$ accuracy for DeepSeek), indicating that it is overly penalizing. In contrast, the Liberal Strategy (orange bars) achieves the highest alignment metrics, particularly for GPT-5 (F1-Score $\approx 79.6\%$). However, we adopt Majority Voting for our main leaderboard to balance sensitivity and specificity, mitigating the risk of false positives inherent in the Liberal approach.

\subsection{discussion}

The empirical results from \textbf{MedDialogRubrics} reveal critical insights into the current state of medical AI and the validity of simulated benchmarks.
\paragraph{The "Inquiry Deficit" in LLMs.}A central finding from Figure~\ref{fig:comparison} is that an increase in context length does not strictly guaranty better diagnostic reasoning. Although prompt engineering techniques often focus on context window utilization, our results suggest that the bottleneck lies in active inquiry planning. Models like Qwen3-235B-A22B-Instruct-2507 and DeepSeek-r1 plateau early, suggesting that they do not dynamically update their differential diagnosis to ask the next most relevant question. Only Gemini-2.5-pro demonstrated the ability to "close the loop" effectively within a standard clinical time frame (8-12 turns).
\paragraph{Benchmarking Beyond Static QA.}The dynamic nature of our results validates the necessity of multi-turn evaluation. In static benchmarks (e.g. MedQA), differences between models like GPT-5 and Qwen3-235B-A22B-Instruct-2507 might be marginal. However, our temporal analysis (Figure~\ref{fig:comparison}) exposes a significant behavioral gap—up to a 20\% difference in rubric coverage—that static snapshots would obscure. This confirms that \textbf{MedDialogRubrics} captures the clinical reasoning process, not just the outcome.
\label{reliability}
\paragraph{Reliability of Automated Evaluation.}The high alignment scores in Figure~\ref{fig:autoevaluation} (Acc $>76\%$) between our automated pipeline and human experts are crucial. They demonstrate that by employing a multi-agent judging system with voting ensembles, we can scale medical evaluation without the prohibitive cost of continuous expert annotation. The "Liberal Strategy" shows high agreement for GPT-5 specifically, suggesting that stronger models may generate more nuanced answers that are harder for a strict "Unanimous" judge to validate, but are correctly recognized by a more flexible aggregation strategy.

\section{Conclusions}
We present MedDialogRubrics, a benchmark and evaluation framework for rigorously assessing the multi-turn inquiry abilities of medical LLMs. Unlike previous work that focused on single-turn QA or final diagnosis accuracy, our framework targets fine-grained, human-aligned evaluation of the diagnostic process. Using 5,200 synthetic patient cases and more than 60,000 expert-refined rubric criteria, MedDialogRubrics assesses not only diagnostic correctness but also the completeness, logic, and effectiveness of information gathering.
A key feature is the use of Evidence-Based Medicine (EBM) guidelines to define specific "must-ask" questions, exposing capability gaps that aggregate metrics miss and separating conversational fluency from clinical adequacy. A dynamic guidance mechanism reduces hallucinations during data generation, keeping evaluations clinically plausible and coherent.
Experiments show that even state-of-the-art LLMs struggle under these standards, particularly in strategic information seeking and long-context management. Our results indicate that improving medical conversational AI will require advances in dialogue management architectures, not just incremental tuning of the base-model. MedDialogRubrics, therefore, offers a standardized and challenging platform for reproducible research to advance safer, more comprehensive, and clinically effective doctor agents.

\bibliographystyle{unsrt}  
\bibliography{references}  

\appendix

\section{Supplementary Figures and Tables}
\label{sec:appendix}

\begin{table*}[t]
\centering
\setlength{\tabcolsep}{10pt}
\renewcommand{\arraystretch}{1.25}

\newcolumntype{L}[1]{>{\raggedright\arraybackslash}p{#1}}

\begin{tabular}{L{0.2\textwidth} L{0.35\textwidth} L{0.35\textwidth}}
\toprule
\textbf{Category} & \textbf{Description} & \textbf{Example} \\
\midrule

\makecell[tl]{\textbf{Symptom}\\\textbf{Characterization}}&
Systematically describe the main symptom, including onset and time course, location/radiation, quality, severity, triggers/relievers, associated symptoms, and response to any self-care. &
"When did it start—sudden or gradual?"; "Where exactly is it? Does it spread anywhere?"; "What does it feel like (sharp, dull, burning, cramping)? Rate 0–10."; "What makes it worse or better (activity, food, position, rest, medication)?" \\
\midrule

\makecell[tl]{\textbf{Urgency/Triage}\\\textbf{Assessment}} &
Screen for warning signs that suggest severe disease or clinical instability, and determine whether urgent evaluation, emergency referral, or immediate testing is needed. &
"Any chest tightness/pain, sweating, shortness of breath, palpitations, or pain radiating to the left arm/jaw?"; "Any confusion, fainting, slurred speech, weakness/numbness, seizures?"; "High fever that won’t come down, chills/rigors, purple rash/petechiae?" \\
\midrule

\makecell[tl]{\textbf{Exploration of}\\\textbf{Differential}\\\textbf{Diagnostics}} &
Ask discriminating questions that help confirm or rule out the most likely and the most dangerous alternative diagnoses relevant to the presentation. &
(Abdominal pain) "Is the pain localized? Any rebound pain or pain that moves? Worse after meals or on an empty stomach?"; (Fever/cough) "Any sputum—color/amount? Pleuritic chest pain? Sick contacts? Vaccination status?"; (Dizziness) "Is it spinning vs lightheadedness on standing? Any tinnitus or hearing loss?" \\
\midrule

\makecell[tl]{\textbf{Medical history \&}\\\textbf{Individual Risk }\\\textbf{Context}} &
Review medical background and risk factors that influence diagnosis, treatment choices, and medication safety (conditions, surgeries, allergies, medications, family history, reproductive status when relevant). &
"Any history of hypertension, diabetes, coronary disease, asthma, liver/kidney disease?"; "Any surgeries or hospitalizations? Recent tests or imaging?"; "Any medication/food allergies? Ever had severe reactions?"; "What medications are you taking (dose/frequency)? Any recent starts/stops?" \\
\midrule

\makecell[tl]{\textbf{Social and}\\\textbf{Lifestyle Factors}} &
Identify exposures and lifestyle factors that change disease likelihood or affect prevention and management (occupation, travel, contacts, substances, diet, sexual and environmental exposures as appropriate). &
"What do you do for work? Any exposure to dust, chemicals, noise, radiation, animals?"; "Any recent travel/camping? Any mosquito/tick bites?"; "Any close contacts with fever/cough/diarrhea? Recent hospital/large gatherings?" \\
\midrule

\makecell[tl]{\textbf{Functional}\\\textbf{Impact}} &
Assess how symptoms affect daily life and functioning, and clarify the patient’s concerns, expectations, and preferred care goals to support shared decisions. &
"How is this affecting work/school—have you needed time off?"; "Any impact on sleep, appetite, energy? Trouble walking or climbing stairs?"; "What bothers you most? What are you most worried about?" \\
\bottomrule
\end{tabular}
\caption{Key-rubric categories with descriptions and examples.}
\label{tab:rubric_criteria}
\end{table*}

\appendix
\begin{figure}[!htbp]
    \centering
    \includegraphics[width=1\linewidth]{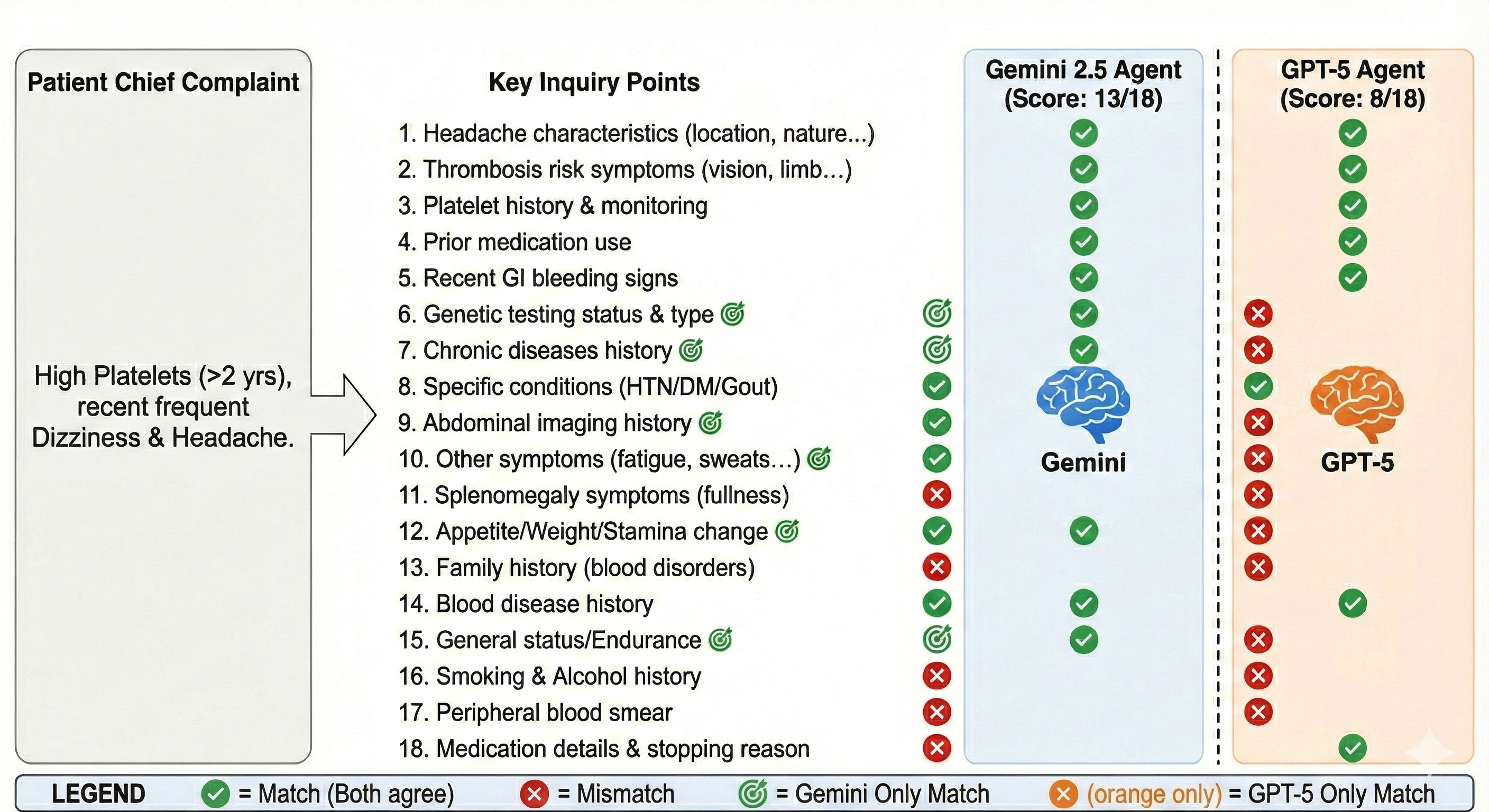}
    \caption{An Example of the Key Rubric matching between Gemini-2.5-pro and Gpt-5}
    \label{fig:example}
\end{figure}

\end{document}